# Comparative Evaluation of Machine Learning Algorithms for Affective State Recognition from Children's Drawings

Aura-Loredana Dan (Popescu), Bucharest, Romania,
apopescu1705@stud.acs.upb.ro,
https://orcid.org/0000-0003-4786-3455

*Abstract*—**Autism spectrum disorder (ASD) represents a neurodevelopmental condition characterized by difficulties in expressing emotions and communication, particularly during early childhood. Understanding the affective state of children at an early age remains challenging, as conventional assessment methods are often intrusive, subjective, or difficult to apply consistently. This paper builds upon previous work on affective state recognition from children's drawings by presenting a comparative evaluation of machine learning models for emotion classification. Three deep learning architectures—MobileNet, EfficientNet, and VGG16—are evaluated within a unified experimental framework to analyze classification performance, robustness, and computational efficiency. The models are trained using transfer learning on a dataset of children's drawings annotated with emotional labels provided by psychological experts. The results highlight important trade-offs between lightweight and deeper architectures when applied to drawing-based affective computing tasks, particularly in mobile and real-time application contexts.**

*Index Terms*—**Affective computing, autism spectrum disorder, children's drawings, deep learning, emotion recognition, machine learning, transfer learning.**

## 1. INTRODUCTION

Autism spectrum disorder is a neurodevelopmental condition characterized by persistent difficulties in social communication, emotional expression, and speech development, with symptoms typically emerging in early childhood [1]. Speech and language impairments are among the most common challenges associated with autism spectrum disorder and often limit a child's ability to express internal emotional states through conventional verbal communication [2]. Early identification and intervention are essential, as timely support can significantly improve long-term developmental outcomes [3][6]. However traditional assessment methods frequently rely on direct interaction, clinical observation, or caregiver reports, which may be intrusive, subjective, or difficult to apply consistently [7]. Consequently, alternative, non-invasive approaches for understanding children's emotional states are increasingly being explored. Expressive activities such as drawing provide a natural and accessible medium for emotional communication, particularly for children with limited speech abilities [9]. Recent advances in machine learning and affective computing enable the objective analysis of such expressive data, opening new possibilities for digital tools that support emotional understanding and communication in children with neurodevelopmental challenges [8][5][12].

## 2. BACKGROUND AND RELATED WORK ON DRAWING-BASED EMOTION ANALYSIS

Children's drawings have long been studied as a meaningful medium for emotional expression, particularly in early childhood, when verbal communication skills are still developing [6][7]. For children with neurodevelopmental conditions such as autism spectrum disorder and speech impairments, drawings often serve as an alternative channel for expressing internal emotional states that may be difficult to communicate verbally [9]. Psychological studies have shown that elements such as color usage, spatial organization, shape selection, and drawing content can reflect affective and cognitive processes in young children.

In recent years, advances in machine learning and affective computing have enabled the automated analysis of visual artifacts, including children's drawings, for emotion recognition tasks [13]. Early approaches relied on handcrafted features combined with classical classifiers, while more recent work has focused on deep learning architectures capable of learning hierarchical visual representations directly from image data [8][10][11]. Convolutional neural networks have been successfully applied to drawing-based emotion analysis, demonstrating promising results in recognizing affective patterns across different developmental stages.

Prior research has also explored the feasibility of integrating such models into digital and mobile applications, highlighting the importance of computational efficiency, robustness, and interpretability [10]. Lightweight architectures are particularly relevant for real-time and mobile scenarios, where resource constraints are critical [14]. However, deeper models may capture more complex visual features, potentially improving classification performance at the cost of increased computational demands. This evolving research landscape motivates a systematic comparison of deep learning architectures to better understand their suitability for drawing-based affective state recognition in children.

## 3. MACHINE LEARNING MODELS AND EXPERIMENTAL FRAMEWORK

This work builds upon the authors' previous research on affective state recognition from children's drawings, where convolutional neural networks were employed to classify emotional states using visual features extracted from free-form drawings [2]. In that study, an initial dataset and experimental framework demonstrated the feasibility of using machine learning techniques to support non-invasive emotional assessment in children with neurodevelopmental challenges [8][9][10]. The present work extends this line of research by introducing an expanded dataset, a refined experimental protocol, and a comparative evaluation of multiple deep learning architectures. All models were trained using the same experimental protocol, including identical training–testing splits (75% training, 25% testing), batch size, early stopping criteria, and optimization strategy. Model performance was evaluated using accuracy, precision, recall, F1-score, and confusion matrices. In addition, training convergence behavior and inference efficiency were analyzed to assess practical deployment feasibility.

*3.1. Emotional State Definition and Dataset Description*

The dataset used in this study consists of 1,472 children's drawings, collected and annotated with expert input from psychological specialists. The drawings are categorized into five dominant emotional states: happy, sad, angry, fear, and insecure. These states were selected based on their frequent occurrence in early childhood emotional expression and their relevance in the psychological assessment of children with autism spectrum disorder and related neurodevelopmental conditions [4]. Prior studies in child psychology indicate that these affective states are commonly reflected through visual elements such as color usage, stroke intensity, spatial organization, and symbolic content in drawings [6][7].

Rather than attempting fine-grained or overlapping emotional labels, the dataset focuses on dominant affective states to reduce subjectivity and improve annotation consistency [5][10]. This choice aligns with clinical practice, where primary emotional tendencies are often more informative than subtle emotional variations, particularly in children with limited verbal communication abilities.

As illustrated in Fig. 1, the dataset exhibits moderate class imbalance, with the happy class being the most represented. To address this, data augmentation techniques were applied uniformly across classes during training to improve generalization and robustness.

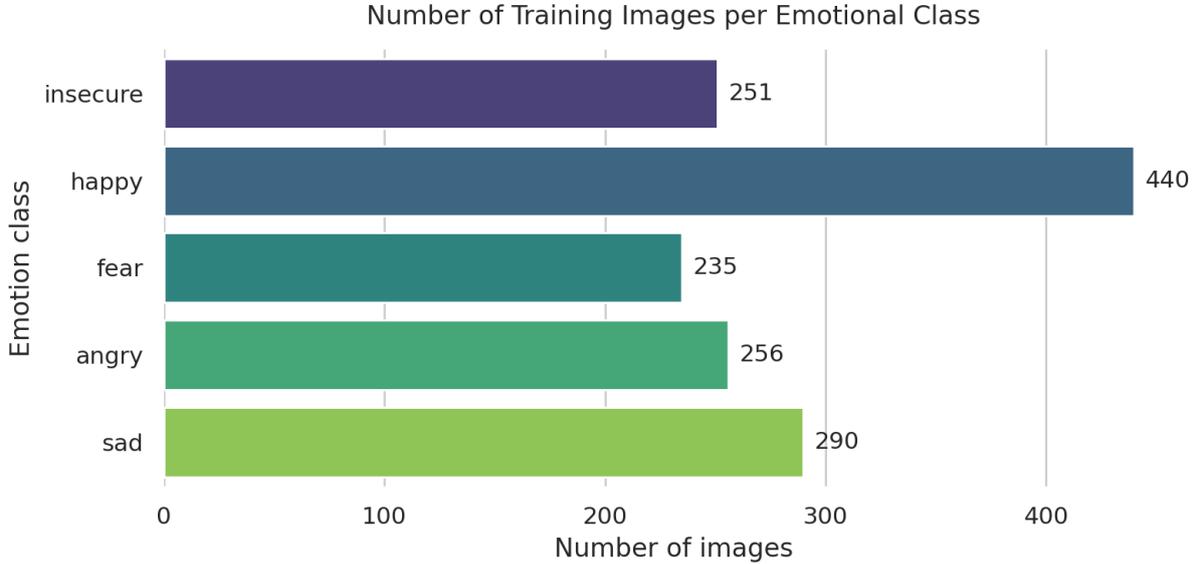

**Fig. 1** Number of Training images per Emotional class

*3.2. Data Preprocessing and Augmentation*

All drawings were resized to a fixed resolution of 224 × 224 pixels and converted to RGB (Red, Green, and Blue) format to match the input requirements of the pretrained convolutional neural networks. Pixel intensities were preserved in the original range and normalized internally using architecture-specific preprocessing functions.

Given the moderate dataset size and the presence of class imbalance, data augmentation was employed during training to improve generalization and reduce overfitting. The augmentation strategy included small geometric transformations that preserve the semantic content of children's drawings, namely: random rotations (±15°), slight zoom variations (±5%), and horizontal and vertical translations (±5%). Horizontal flipping was intentionally excluded to avoid altering spatial cues that may be emotionally meaningful in children's drawings. In Figure 2, is presented the drawing with augmentation. Augmentation was applied only to the training set and generated on-the-fly during model optimization.

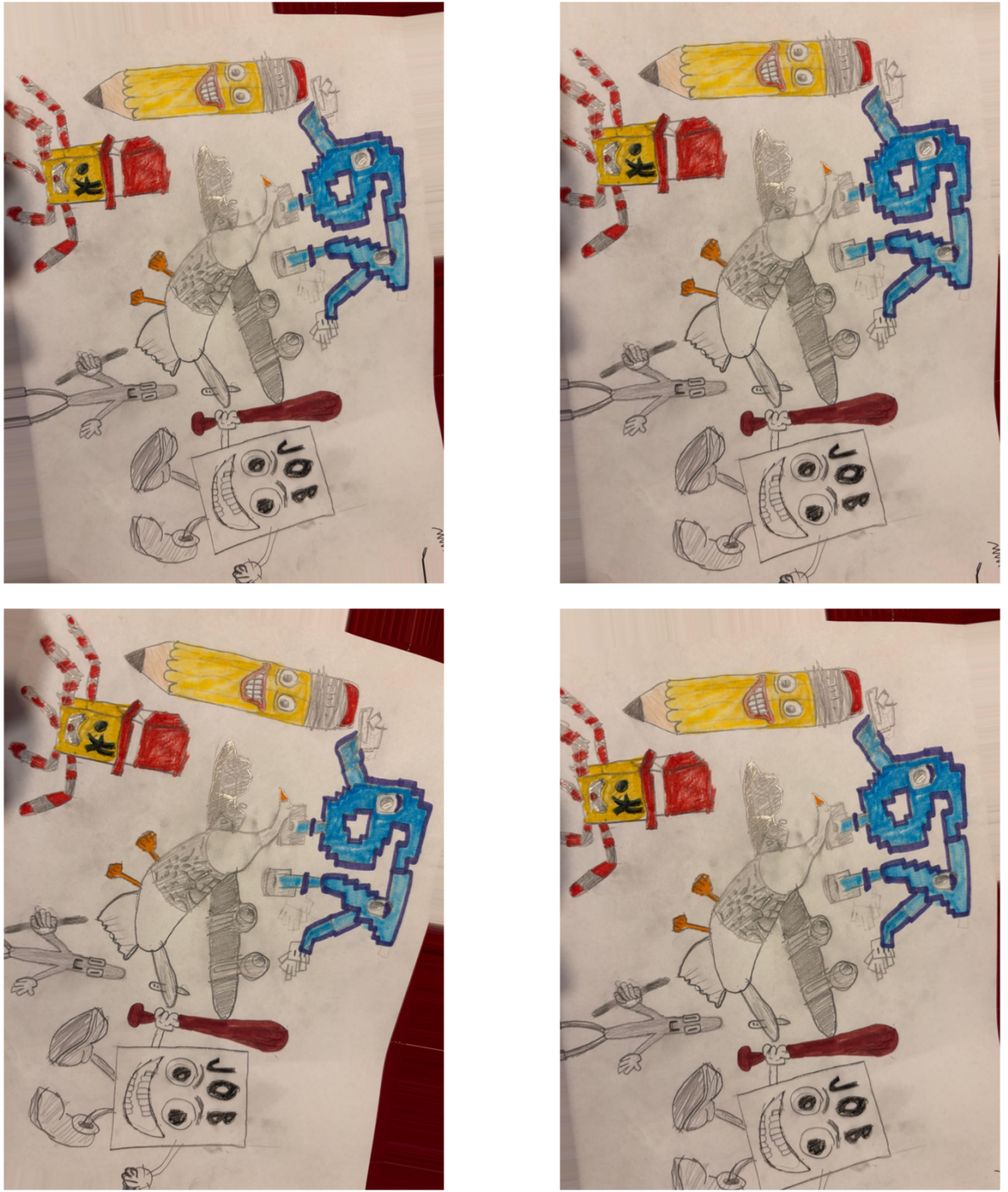

**Fig. 2** Number of Training images per Emotional class

*3.3. MobileNet Architecture*

MobileNetV2 [17] was selected as a lightweight architecture optimized for mobile and real-time applications. The pretrained convolutional base, initialized with ImageNet weights, was used as a fixed feature extractor in the initial training phase. A custom classification head was added, consisting of a global average pooling layer, followed by two fully connected layers with dropout regularization and a softmax output layer corresponding to the five emotional classes.

*3.4. Vgg16 Architecture*

VGG16 [16][19] was used as a deeper baseline architecture to evaluate the impact of increased representational capacity on emotion recognition performance. Similar to MobileNet, the pretrained VGG16 convolutional layers were initialized with ImageNet weights and kept frozen during training. To mitigate overfitting associated with deep architectures and limited data, a lightweight classification head was employed, replacing traditional fully connected layers with global average pooling and dropout regularization.

While VGG16 provides strong feature extraction capabilities, its large parameter count and computational demands make it less suitable for mobile deployment, serving instead as a performance-oriented comparison model.

*3.5. EfficientNet Architecture*

EfficientNetB0 [18[15]] was included due to its compound scaling strategy, which jointly optimizes network depth, width, and resolution. This architecture has demonstrated strong performance across vision tasks while maintaining computational efficiency. As with the other models, the EfficientNet backbone was initialized with pretrained ImageNet [20] weights and combined with a lightweight classification head identical in structure to those used for MobileNet and VGG16 to ensure a fair comparison.

EfficientNet achieved the most favorable balance between classification accuracy and computational efficiency among the evaluated models, making it particularly suitable for affective computing applications that require both robustness and scalability.

## 4. EXPERIMENTAL RESULTS AND PERFORMANCE EVALUATION

*4.1. Overall Classification Performance*

This section presents the quantitative evaluation of the proposed models for affective state recognition from children's drawings. Performance was assessed using classification accuracy, loss values, and confusion matrix analysis on the held-out test set. Three transfer learning architectures—MobileNet, VGG16, and EfficientNet—were evaluated under identical training conditions to ensure a fair comparison.

Figure 5 illustrates the training and validation accuracy curves obtained over 20 training epochs for the EfficientNet model. The training accuracy increases rapidly during the initial epochs and continues to improve steadily, reaching a high convergence level close to the end of training. In contrast, the validation accuracy stabilizes at approximately 62.77%, indicating that EfficientNet is able to learn discriminative representations while maintaining comparatively good generalization performance. The gap between training and validation curves suggests mild overfitting; however, the validation trend remains stable, demonstrating robustness on unseen data.

Figure 4 presents the accuracy curves for the MobileNet architecture. The training accuracy increases consistently across epochs, but at a slower rate than EfficientNet, while the validation accuracy plateaus around 59.24%. Compared to EfficientNet, MobileNet exhibits a larger discrepancy between training and validation performance, reflecting its more lightweight architecture and reduced representational capacity. Nevertheless, the relatively stable validation curve confirms that MobileNet remains suitable for resource-constrained and mobile deployment scenarios.

Figure 3 shows the accuracy evolution for the VGG16 model. Both training and validation accuracies remain significantly lower than those of EfficientNet and MobileNet, with validation accuracy reaching only 46.2% after 20 epochs. The limited improvement across epochs suggests that VGG16 struggles to adapt effectively to the drawing-based emotional classification task under the given dataset size and training configuration. This behavior is likely due to the model's high parameter count, which increases the risk of underfitting or inefficient feature learning when applied to small and specialized datasets.

Overall, the accuracy curves demonstrate that EfficientNet achieves the best balance between convergence speed,final performance, followed by MobileNet, while VGG16 yields the weakest results for this task.

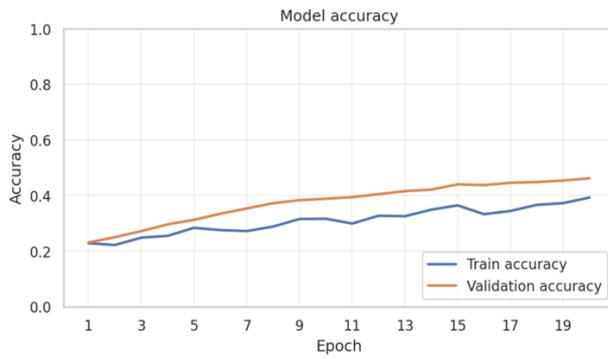

**Fig. 3** VGG16 model accuracy

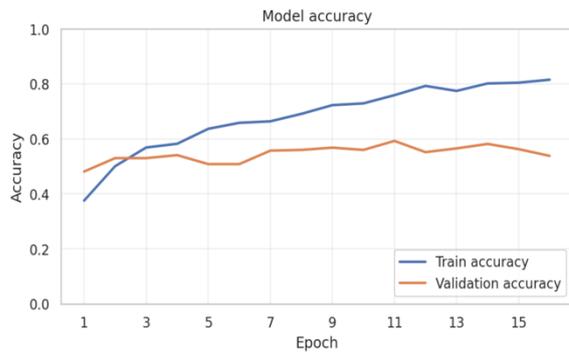

**Fig. 4** MobileNet model accuracy

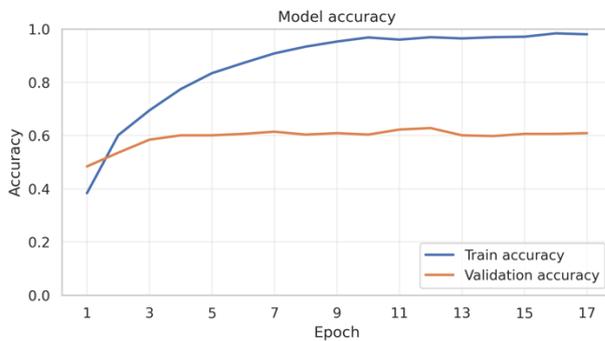

**Fig. 5** EfficientNet model accuracy

Among the evaluated models, EfficientNet achieved the highest classification accuracy of 62.77%, with a corresponding loss value of 1.8688, demonstrating superior generalization capability on the test data. MobileNet achieved an accuracy of 59.24% with a lower loss of 1.1821, reflecting its efficient learning behavior and suitability for lightweight deployment scenarios. In contrast, VGG16 obtained a significantly lower accuracy of 46.2% and a loss of 1.7367, indicating limited effectiveness for this task given the dataset size and complexity.

These results confirm that architectures specifically optimized for efficiency and balanced scaling, such as EfficientNet, are better suited for affective computing tasks involving children's drawings, where both subtle visual patterns and generalization are critical.

Beyond the quantitative evaluation, the findings of this study have direct practical relevance within the context of the PandaSays [21] application, a digital assistive tool designed to support emotional expression and communication in children diagnosed with autism. The application leverages children's free-form drawings as a natural interaction modality, aiming to provide caregivers and specialists with insights into the child's affective state. The superior performance of EfficientNet observed in this work supports its suitability for integration into PandaSays, where robust generalization and efficient inference are essential. By enabling more accurate emotion recognition from children's drawings, the proposed approach contributes to the development of non-invasive, AI (Artificial Intelligence)-assisted tools that facilitate emotional awareness and communication in real-world therapeutic and educational settings.

*4.2. Confusion Matrix Analysis*

To further analyze model behavior beyond aggregate accuracy, confusion matrices were computed for each architecture, as illustrated in Figures 6–8. The confusion matrix of EfficientNet shows improved class discrimination across all five emotional states, with particularly strong performance in recognizing happy and fear categories. Misclassifications primarily occurred between emotionally adjacent classes, such as sad and insecure, which share overlapping visual characteristics in children's drawings.

MobileNet demonstrates comparable performance, though with slightly higher confusion between angry and sad states. This behavior reflects the trade-off between reduced model complexity and representational capacity. Nevertheless, MobileNet maintains stable performance across all classes, reinforcing its applicability in mobile and real-time systems.

VGG16 exhibits the highest degree of misclassification across all emotional categories. The confusion matrix reveals frequent incorrect predictions, particularly for minority classes, suggesting overfitting and insufficient adaptation of deep features to the affective drawing domain.

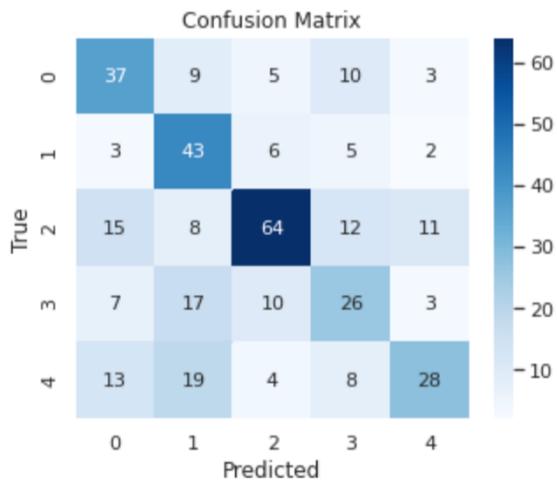

**Fig. 6** MobileNet confusion matrix

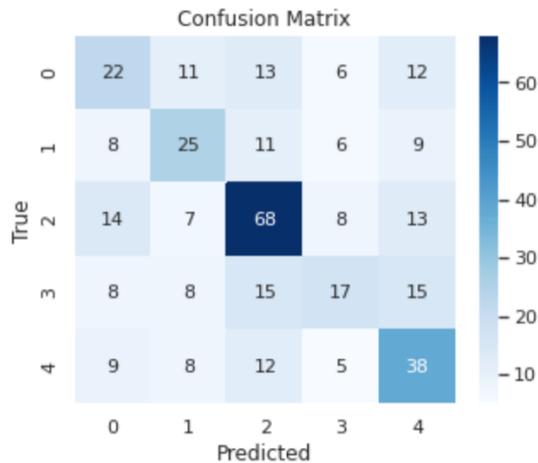

**Fig. 7** VGG16 confusion matrix

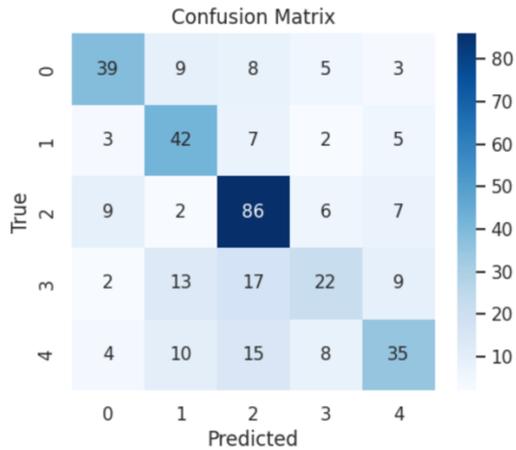

**Fig. 8** EfficientNet confusion matrix

## 5. DISCUSSION

This section discusses the comparative performance of the evaluated deep learning architectures and analyzes the factors influencing their effectiveness in recognizing affective states from children's drawings. The results demonstrate that EfficientNet achieves the highest classification accuracy (62.77%), outperforming MobileNet (59.24%) and VGG16 (46.20%), while maintaining a favorable balance between model complexity and computational efficiency.

The superior performance of EfficientNet can be attributed to its compound scaling strategy, which jointly optimizes network depth, width, and input resolution. This design enables EfficientNet to capture richer visual representations from children's drawings without a proportional increase in computational cost. Compared to the results reported in the authors' previous work on drawing-based emotion recognition, where lighter architectures were primarily explored, the current findings indicate a consistent improvement in accuracy when EfficientNet is employed under a unified training and augmentation framework. This suggests that this neural network is particularly well suited for affective computing tasks involving abstract and non-photorealistic visual inputs.

MobileNet demonstrates competitive performance while remaining computationally efficient, confirming its suitability for mobile and real-time applications. Although its accuracy is slightly lower than that of EfficientNet, MobileNet benefits from a significantly reduced parameter count and faster inference, which are critical considerations for deployment in assistive technologies targeting children. In contrast, VGG16 exhibits the lowest performance among the evaluated models. Its deep and parameter-heavy architecture appears less effective for the relatively small and heterogeneous dataset of children's drawings, likely due to overfitting and limited generalization capability.

An additional challenge observed across all models is the confusion between emotionally adjacent classes, such as sad and insecure, as reflected in the confusion matrix analysis. These ambiguities highlight the intrinsic subjectivity of emotional expression in children's drawings and reinforce the importance of psychological expertise in dataset annotation. Overall, the findings support the use of efficient deep learning architectures for non-invasive emotional state recognition and provide evidence that carefully balanced models can outperform deeper networks in this domain.

## 6. CONCLUSION

This paper presented a comparative study of deep learning architectures for affective state recognition from children's drawings, with a particular focus on applications for children with autism spectrum disorder. By extending previous research in this domain, a unified experimental framework was employed to evaluate MobileNet, VGG16, and EfficientNet using a dataset of 1,472 annotated drawings representing five dominant emotional states. The results demonstrate that EfficientNet achieves the best overall performance, offering an effective balance between classification accuracy and computational efficiency, while MobileNet remains a strong candidate for deployment in mobile and resource-constrained environments. In contrast, VGG16 showed limited generalization capability for this task, highlighting the importance of architecture selection when working with abstract and expressive visual data.

Beyond quantitative performance, the analysis revealed consistent challenges related to the overlap between emotionally adjacent classes, reflecting the subjective and nuanced nature of emotional expression in children's drawings. These findings emphasize the need for psychologically informed labeling and model interpretability in affective computing systems.

The outcomes of this work support the feasibility of non-invasive, machine learning–based tools for emotional state recognition and their potential integration into digital assistive technologies aimed at enhancing emotional awareness and communication. Future work will focus on expanding the dataset, exploring multimodal approaches that combine visual and behavioral cues, and investigating personalized and longitudinal models to better capture individual emotional development patterns in children with neurodevelopmental challenges.

**Statements and Declarations**

*Availability of data and materials*
The datasets generated and/or analyzed during the current study are not publicly available but are available from the corresponding author on reasonable request.

The data was fully anonymized, collected with informed consent from parents or legal guardians, analyzed retrospectively without any direct intervention involving the children, and posed no risk of identification or harm to participants.

*Competing interests*
The author declares that there are no competing interests.

*Funding*
This research received no external funding.

*Authors' contributions*
The author conceived the study, designed the methodology, performed the experiments, analyzed the results, and wrote the manuscript.

*Acknowledgements*
The author thanks the clinical psychologists who contributed their expertise to the annotation and interpretation of children's drawings used in this study. Their professional insight was essential in defining the emotional categories and validating the labelling of the drawings, ensuring that the dataset reflects psychologically meaningful affective states.

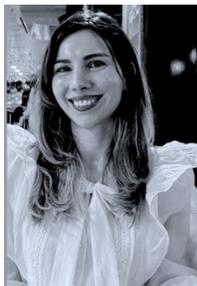

**Aura-Loredana Dan,** received the Ph.D. degree in computer science from the National University of Science and Technology POLITEHNICA Bucharest, Bucharest, Romania, within the Computer Science Department. Her doctoral research focused on machine learning and affective computing for children with neurodevelopmental disorders, with particular emphasis on the automatic interpretation of emotional states from children's drawings. Her research has resulted in several peer-reviewed publications, including "Neural Network–Based Solutions for Predicting the Affective State of Children with Autism" (2021 23rd International Conference on Control Systems and Computer Science (CSCS)), "Machine Learning–Based Solutions for Predicting the Affective State of Children with Autism" (2020 International Conference on e-Health and Bioengineering (EHB)), and "An IoT- and AI-Based Application for the Automatic Interpretation of the Affective State of Children Diagnosed with Autism" (Sensors, 2022).

Dr. Dan is currently engaged in applied artificial intelligence research and software development, with experience in designing and deploying machine learning models for real-world applications. Her work focuses on bridging academic research and practical implementation of intelligent systems aimed at supporting emotional understanding, communication, and assistive technologies for children with neurodevelopmental challenges.